\documentclass[letterpaper]{article} % DO NOT CHANGE THIS
\usepackage[hyphens]{url}  % DO NOT CHANGE THIS
\usepackage{graphicx} % DO NOT CHANGE THIS
\usepackage{natbib}  % DO NOT CHANGE THIS AND DO NOT ADD ANY OPTIONS TO IT
\usepackage{caption} % DO NOT CHANGE THIS AND DO NOT ADD ANY OPTIONS TO IT
\usepackage{algorithm}
\usepackage{algorithmic}
\usepackage{amsmath}
\usepackage{multirow}
\usepackage{amssymb}
\usepackage{tcolorbox}
\usepackage{newfloat}
\usepackage{listings}
\DeclareCaptionStyle{ruled}{labelfont=normalfont,labelsep=colon,strut=off} % DO NOT CHANGE THIS
\floatstyle{ruled}
\newfloat{listing}{tb}{lst}{}
\floatname{listing}{Listing}

\usepackage{booktabs}

\title{Breaking the Long Horizon: Short-Horizon Process Supervision for \\ Long-Form Generation in LLMs}
\author{Heyang Xu, HanWen Hao, Qianren Mao, Zhuo Liu, Qili Zhang, Sheng Yang, Jianxin Li}

\begin{document}

\maketitle

\begin{abstract}
Long-form generation is increasingly important for agentic LLM systems, yet current reinforcement learning methods typically optimize an entire document with a single outcome-level reward. This delayed and coarse-grained supervision makes it difficult to localize errors and assign credit across semantically distinct parts of the output. To address this, we propose a short-horizon framework that reformulates long-form generation through semantic decomposition into segment-level decisions. Each segment is generated with prefix-conditioned context, allowing cross-segment coherence to be preserved while keeping supervision local. The generated segments are then scored with rubric-based rewards for adherence, coherence, and relevance, enabling explicit local supervision at the segment level. These local rewards are used to compute step-wise advantages, yielding localized policy updates. To isolate the effect of supervision granularity, we instantiate both outcome-level supervision and our short-horizon process reward method (SH-PRM) under the same GRPO framework. Across multiple LLMs and two long-form generation benchmarks, SH-PRM consistently outperforms outcome-level supervision and strong inference-time baselines under the same optimizer and comparable budgets. Further analyses suggest that these gains primarily arise from improved credit assignment. Overall, our results suggest that aligning supervision granularity with the semantic structure of long-form generation is an effective design principle for long-form LLM training.
\end{abstract}


\section{Introduction}
% 说明任务
Long-form generation refers to the task of generating lengthy, coherent documents that demand global planning and structured content construction~\cite{liu2024lost, lialin2023scaling, he2026stable}. Compared with short-text dialogue, it places substantially stronger demands on both breadth and depth ~\cite{huang2025survey, wei2022chain}, and therefore necessitates complex hierarchical planning capabilities~\cite{xiong2025outlining}. This structured, multi-stage nature makes accurate credit assignment particularly challenging in long-form generation~\cite{zhang2026reasoning, guo2024deepseek, yao2023tree, sun2022recitation}.

% 当前方法及问题
The prevailing training paradigm for such tasks typically relies on reinforcement learning with Outcome Reward Models (ORM)~\cite{huang2026long, ouyang2022training, ziegler2019fine, askell2021general}. 
However, recent work has highlighted important limitations of ORM in long-form generation(LFG)~\cite{tan2026hindsight, lightman2023let,bai2022constitutional,Longrm25}. In particular, the sparse terminal feedback provided by ORM makes it difficult for the model to identify and correct the specific paragraphs responsible for a low score during on-policy learning~\cite{deng2026densegrpo, VinePPO24,Reward25}.

\begin{figure*}[t]
    \centering
    \includegraphics[width=1.0\textwidth]{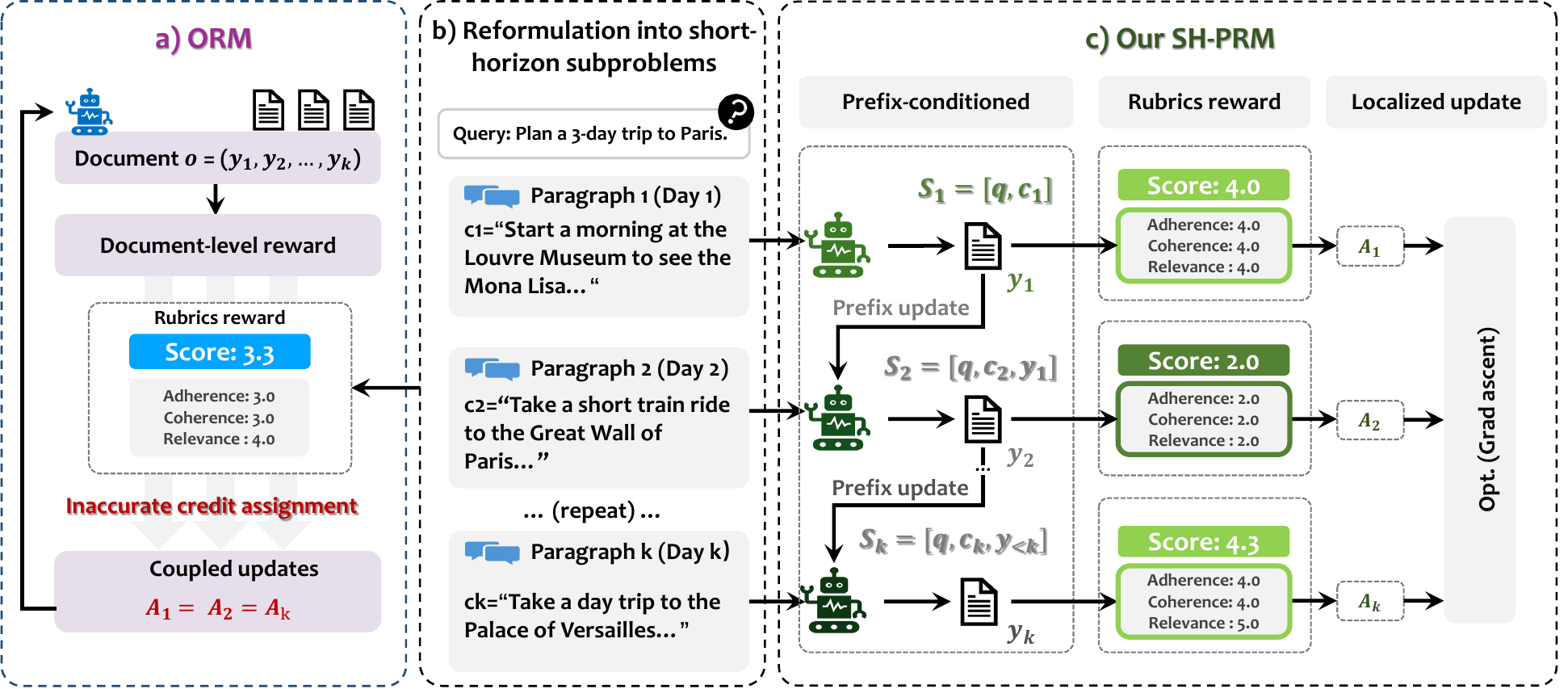}
    \caption{\textbf{Document-level versus segment-level supervision for long-form generation.}
  Under ORM (left), the full document receives a single document-level reward, and the resulting advantage is shared across segments, coupling their updates and potentially obscuring local errors.
  Under our SH-PRM (right), generation is decomposed into segments with prefix-conditioned context, each of which receives a local rubric-based reward and a localized update. 
  This change in supervision granularity enables more precise credit assignment by penalizing problematic paragraphs while preserving credit for strong ones.}
    % \caption{\textbf{Conceptual Comparison of Reward Mechanisms: Outcome Reward Model (ORM) vs. Process Reward Model (PRM).}
    % The \textbf{Left Panel} illustrates the \textit{Reward Collapse} phenomenon in ORM. A single, holistic reward (Score: 3.3) for the entire output can mask severe errors (e.g., "Great Wall of Paris" hallucination in Paragraph 2), leading to imprecise learning and potentially encouraging "confident lying."
    % The \textbf{Right Panel} depicts our PRM, which provides dense, segment-level feedback using multi-dimensional rubrics (Adherence, Coherence, Relevance). This fine-grained scoring precisely identifies and penalizes errors at the paragraph level (e.g., Paragraph 2 scores 2.0), while correctly crediting accurate segments. This mechanism effectively solves the credit assignment problem, facilitating \textit{Stable Convergence} and enabling the model to learn efficiently by "fixing errors while retaining correct parts."}
    \label{fig:main_concept}
\end{figure*}

% 我们的解决
To address this issue, our key idea is that long-form generation is inherently compositional, and training should therefore align supervision granularity with semantic segments rather than treat the entire document as a single optimization unit, as illustrated in Figure~\ref{fig:main_concept}. Based on this idea, we develop a short-horizon, process-supervised training framework for long-form generation. At its core, we reformulate the problem as a sequence of short-horizon semantic subproblems to unlock fine-grained flexibility, establishing a Short-Horizon Generation Paradigm. To support this paradigm, we design a novel short-horizon process reward method (SH-PRM) for model training. 
By providing fine-grained rewards to individual semantic segments, this SH-PRM enables each unit to be optimized according to its own quality while reducing cross-segment interference.

% 理论部分-三句话
% 说明有理论解释。
% 说明markov分解后，策略梯度中存在credit assignment interference。
% 说明b条件，消除了方差干扰，证明我们prm的价值。
Distinct from purely empirical work, we provide a formal analysis of the credit assignment problem in long-form generation. By formulating the generation process as a Markov Decision Process, we mathematically decompose the policy gradient to reveal that outcome-level rewards suffer from severe cross-segment credit assignment interference. 
We further disentangle this effect from pure variance reduction using a controlled baseline, showing that the weakness of outcome-level supervision is not merely a statistical noise issue, but a credit-assignment problem rooted in reward granularity~\cite{Optimization18}.

Experiments across Qwen~\cite{yang2025qwen3} and Llama~\cite{grattafiori2024llama} models on DeepResearch Bench~\cite{deep25} and WritingBench~\cite{writing25} show consistent gains in both long-form generation quality and sample efficiency. On Qwen3-32B, our method improves the overall score from 35.85 to 38.84, outperforming an outcome-level baseline under the same training budget.

% 一句做了什么。
Therefore, our core contributions are listed as follows:
\begin{itemize}
\item We reformulate long-form generation (LFG) as a sequence of short-horizon semantic subproblems, thereby aligning the granularity of supervision with the inherently compositional structure of long-form writing.
\item We introduce SH-PRM, a short-horizon process reward method that provides segment-level feedback and localized advantage estimation to reduce cross-segment interference during optimization. Through theoretical analysis and a controlled baseline, we disentangle structural credit-assignment effects from pure variance reduction in long-form RL, showing that the gains cannot be explained by variance reduction alone and are closely tied to improved credit assignment.
\item Across Qwen and Llama models on DeepResearch Bench and WritingBench, we show that SH-PRM consistently improves benchmark performance and training efficiency under the same optimizer and comparable budgets.
\end{itemize}

\section{Related Work}
\subsection{Long-form Generation}

As large language models (LLMs) are increasingly used in agentic systems and deep research workflows, long-form generation has become an important capability~\cite{wang2025loongrl}. Prior work often treats long-form generation as a multi-stage process involving planning, information seeking, verification, and final writing~\cite{Deep25zhang}. Existing work has improved this pipeline through retrieval augmentation~\cite{tongyi2025, hu2025step}, outline planning~\cite{SuperWriter25, Re22}, and multi-agent collaboration~\cite{CollabLLM25}. While these approaches are effective for information gathering and organization, they place relatively less emphasis on the final report-writing stage itself. Our work focuses on this stage, namely the generation of long, structured reports from a given query and outline.

\subsection{Reinforcement Learning for LFG}

Reinforcement learning has become a common paradigm for aligning text generation with human preferences and task objectives. Recent efforts have extended such training to long-form generation and related generation settings~\cite{bai2024longalign, wu2023fine, RaR25, Rubricscaff25, Drtulu25, liao2026rlmr}. However, these approaches still commonly assign supervision only at the level of the full document, providing a single outcome reward after the entire output is generated~\cite{Rethinking25}. As outputs become longer, this feedback becomes increasingly sparse, since many paragraphs or thousands of tokens are optimized against one scalar signal. This makes it difficult to attribute success or failure to specific parts of the document, creating a substantial challenge in credit assignment.

\subsection{Process Supervision and Reward Decomposition}
Process supervision has shown promise in reasoning tasks, where intermediate steps are often more structured and can be evaluated against relatively explicit criteria~\cite{towards25, rprm25}. In contrast, long-form generation is more open-ended: intermediate segments can have many valid realizations and typically do not admit unique ground-truth labels~\cite{stiennon2020learning, liu2023ring, huang2026long}. This makes it difficult to transfer process supervision directly from reasoning to long-form writing. The central challenge lies in choosing an appropriate semantic granularity for decomposition~\cite{Reverse25} and in defining local rewards based on segment quality rather than exact correctness~\cite{writing25, Researchrubrics25, judging23, LLMRubric24}. Our work addresses this challenge by introducing process supervision over semantic segments for long-form generation.

\section{A Short-Horizon Framework for LFG}
\label{sec:method}
\subsection{Motivation and Formulation}
\label{sec:method_overview}
% 强调动机
Directly generating an entire document in one pass offers limited flexibility in controlling the generation process. 
Furthermore, the resulting sparse feedback signals make it difficult for models to pinpoint local errors during training.

% 洞察
From chain-of-thought prompting to agentic workflows, decomposing complex tasks into sequential steps has proven a fundamental principle for capable 
language models. 
Unlike mathematical reasoning, where only the final answer tokens carry
direct reward signal, long-form generation is inherently compositional:
the final document is the concatenation of segments, each of which constitutes
a complete and independently assessable semantic unit.
Inspired by the decomposing principle and the compositional nature of long-form text, we reformulate the problem as a sequence of short-horizon semantic subproblems. In our implementation, these semantic units are instantiated at the paragraph level.

% 形式化
To formalize this decomposition, we model the generation process as a finite-horizon Markov Decision Process (MDP). A trajectory is denoted by
\begin{equation}
\tau = (s_1, a_1, \dots, s_{T}, a_{T}) \quad \tau \sim \pi_\theta,
\end{equation}

where $s_t = (x, y_{0:t-1})$ consists of the input instruction $x$ and the generated prefix up to paragraph $t-1$, and $a_t = y_t$ denotes the paragraph generated at step $t$ under policy $\pi_\theta$.

The state transition is defined by:
\begin{equation}
s_{t+1} = f(s_t, a_t) = (x, y_{1:t-1}),  
\end{equation}

generating paragraph $y_t$ appends it to the current prefix. 

% 解决的问题
Following this MDP formulation~\cite{sutton1998reinforcement}, long-form generation can be naturally realized in a segment-by-segment manner, as detailed in Section~\ref{sec:generation}. Yet existing training paradigms still optimize over the entire document, leaving a central question unresolved: how to design an RL training method whose supervision granularity aligns with the semantic structure of the generation process, thereby eliminating inter-segment interference. A detailed discussion of how we address this issue is presented in Section~\ref{sec:step_wise_advantage}, with further analysis in Section~\ref{sec:method_theoretical_analysis}.

% 阅读完了，这部分已经非常紧凑了，没什么可删减的了

\subsection{Short-Horizon Generation and Training}
\subsubsection{Short-Horizon Generation Paradigm}
\label{sec:generation}
Under the MDP formulation, we adopt a prefix-conditioned short-horizon generation paradigm. At step $k$, the model generates a single segment from the state
\begin{equation}
    s_k = [q, c_k, y_{<k}]
\end{equation}
where $q$ is the query, $c_k$ the target section title, and $y_{<k} = (y_1, \ldots, y_{k-1})$ the history prefix.

The action is $a_k = y_k$, with state transition
\begin{equation}
    s_{k+1} = [q, c_{k+1}, y_{<k} \oplus y_k]
\end{equation}
Conditioning on $y_{<k}$ preserves cross-segment coherence while allowing segment-level control.

\subsubsection{Process-Supervised Reinforcement Learning.}
\label{sec:step_wise_advantage}
We align supervision granularity with the semantic unit of generation so that each segment is credited according to its own quality.
\paragraph{Step-wise Advantage Estimation.}
Under the short-horizon paradigm, each segment is an optimization unit and the advantage is estimated at the segment level. We instantiate both ORM and SH-PRM under the shared Group Relative Policy Optimization (GRPO) framework~\citep{shao2024deepseekmath}, so the comparison isolates reward granularity rather than optimizer choice.

Formally, the policy gradient for segment $y_k$ is
\begin{equation}
    g_k = \mathbb{E}\left[\nabla_\theta \log \pi_\theta(y_k | s_k) \cdot r_k\right]
\end{equation}
where $r_k$ is the reward for segment $k$.

\textit{GRPO Instantiation.} In our SH-PRM-GRPO, the step-wise advantage for candidate $i$ at segment $k$ is
\begin{equation}
    A_{i,k} = \frac{r_{i,k} - \mu_k}{\sigma_k + \epsilon}
    \label{eq:step_wise_advantage}
\end{equation}
where $\mu_k$ and $\sigma_k$ are the mean and standard deviation of rewards across the $G$ candidates at segment $k$.

This normalization compares candidates with the same history prefix $s_k$, so $A_{i,k}$ reflects only the quality of the current segment $y_k^{(i)}$ rather than other segments.

\textit{Comparison with ORM-GRPO} Standard ORM-GRPO computes advantage based on trajectory-level rewards:
\begin{equation}
    A_i^{\text{ORM}} = \frac{R(o_i) - \text{mean}(R)}{\text{std}(R)}
\end{equation}
where $R(o_i)$ is the outcome reward for the complete trajectory $o_i$. This assigns the same advantage to all tokens and introduces inter-segment interference, which we formally characterize and analyze in Section~\ref{sec:method_theoretical_analysis}.

\subsubsection{Segment-Level Reward Design.}
\label{sec:multidimensional_rubric}

After decomposing long-form generation into semantic subproblems, we evaluate each segment with a multi-dimensional rubric. The three dimensions are \textit{Adherence}, which measures alignment with the section title and outline; \textit{Coherence}, which measures logical flow and consistency with the history prefix $y_{<k}$; and \textit{Relevance}, which measures whether the segment addresses the core query $q$.

The segment-level reward is defined as
\begin{equation}
    r_{i,k} = \frac{1}{3} \left( s^{\text{adh}}_{i,k} + s^{\text{coh}}_{i,k} + s^{\text{rel}}_{i,k} \right)
    \label{eq:segment_reward}
\end{equation}
where $s^{\text{adh}}_{i,k}$, $s^{\text{coh}}_{i,k}$, and $s^{\text{rel}}_{i,k}$ denote the scores for Adherence, Coherence, and Relevance. Coherence is especially important because it brings cross-segment dependency into the local reward signal.
To keep the comparison with ORM baselines clean, we use a simple linear average rather than elaborate reward shaping.

\paragraph{Training Algorithm.}
\label{sec:algorithm}

Algorithm~\ref{alg:prm_grpo} summarizes the complete SH-PRM training procedure. 

\begin{algorithm}[t!]
\caption{SH-PRM Training}
\label{alg:prm_grpo}
\begin{algorithmic}[1] % [1] 表示显示行号
    \REQUIRE Policy $\pi_\theta$, SH-PRM $\mathcal{R}$, group size $G$
    \FOR{iteration $= 1, 2, ...$}
        \FOR{each training instance $s_k$ in batch}
            \STATE Sample $G$ candidates: $y_k^{(i)} \sim \pi_{\theta_{\text{old}}}(\cdot|s_k)$ for $i=1,...,G$
            \FOR{$i = 1$ to $G$}
                \STATE $\{a_{i,k}, c_{i,k}, \rho_{i,k}\} \leftarrow \mathcal{R}(y_k^{(i)}, s_k)$
                \STATE $r_{i,k} \leftarrow (a_{i,k} + c_{i,k} + \rho_{i,k}) / 3$
            \ENDFOR
            \STATE $\mu_k \leftarrow \frac{1}{G}\sum_{i=1}^G r_{i,k}$, $\sigma_k \leftarrow \mathrm{std}(\{r_{i,k}\})$
            \FOR{$i = 1$ to $G$}
                \STATE $A_{i,k} \leftarrow (r_{i,k} - \mu_k) / (\sigma_k + \epsilon)$
            \ENDFOR
            \STATE Compute GRPO loss using $\{A_{i,k}\}$ and update $\theta$
        \ENDFOR
    \ENDFOR
\end{algorithmic}
\end{algorithm}

% 结尾
% This design ensures that credit assignment occurs at the segment level, with comparisons among candidates sharing identical input contexts. We present the detailed training objective and implementation specifics in Appendix~\ref{app:training_details}.

\subsection{Theoretical Analysis for SH-PRM}
\label{sec:method_theoretical_analysis}
We analyze why outcome-level supervision is inefficient in long-form generation, and how short-horizon supervision alleviates this problem.
\subsubsection{Structural Flaw: Credit Assignment Inefficiency.}
\label{sec:bias_analysis}
The fundamental distinction between ORM and SH-PRM lies in the granularity of the reward signal. Building upon the MDP formulation established in Section~\ref{sec:method_overview}, ORM generates the entire document in a single pass and receives only a trajectory-level reward:
\begin{equation}
\hat{g}_{\mathrm{ORM}} = \nabla_\theta 
\log \pi_\theta(\tau \mid s) \cdot R(\tau)
\end{equation}

Applying the chain rule of log-likelihood:
\begin{equation}
    \log \pi_\theta(\tau \mid s) = \sum_{t=1}^{T} \log \pi_\theta(a_t \mid s_t).
\end{equation}

Let $R_{-t} = \sum_{k \neq t} r_k$ denote the cross-segment interference, which is the sum of rewards from all segments other than $t$.
Thus, the effective gradient contribution applied to a specific segment $t$ decomposes as:
\begin{equation}
    \hat{g}_{\mathrm{ORM},t} = \nabla_\theta \log \pi_\theta(a_t \mid s_t) \cdot (r_t + R_{-t})
    \label{eq:prm_grad}
\end{equation}

% Thus, the effective gradient contribution applied to a specific segment $t$ decomposes as:
% \begin{equation}
%     \hat{g}_{\mathrm{ORM},t} = \nabla_\theta \log \pi_\theta(a_t \mid s_t) \cdot (r_t + R_{-t})
% \end{equation}
% where $R_{-t} = \sum_{k \neq t} r_k$ aggregates rewards from all segments 
% other than $t$.
\paragraph{Structural Solution: Short-Horizon Supervision.}
SH-PRM effectively circumvents cross-segment interference by decomposing the generation into short-horizon segments, updating the policy step-by-step:
\begin{equation}
    \hat{g}_{\mathrm{SH-PRM},t} = \nabla_\theta \log \pi_\theta(a_t \mid s_t) \cdot r_t
\end{equation}
Each segment is updated using a reward $r_t$, yielding a more localized credit assignment signal. Because the state formulation $s_t$ contains the history prefix $y_{<t}$ and the reward $r_t$ explicitly evaluates \textit{Coherence} and \textit{Adherence} as described in Section~\ref{sec:multidimensional_rubric}, relevant cross-segment dependencies can be reflected in the local reward. This provides a more localized training signal than a noisy global reward. Further derivations are provided in the supplementary material.

\paragraph{Statistical Confound: The Variance Explosion.}
Beyond disrupting the credit assignment structure, cross-segment interference also induces severe structural variance. The variance of the gradient can be decomposed into a variance term and a covariance term:
\begin{equation}
    \mathrm{Var}\left( \sum_{t=1}^T \hat{g}_t \right) = \sum_t \mathrm{Var}\left( \hat{g}_t \right) + 2 \sum_{i<j} \mathrm{Cov}\left( \hat{g}_i, \hat{g}_j \right)
\end{equation}

Since the gradient variance is heavily dominated by the reward variance, we can analyze this through a simplified reward variance decomposition:
\begin{equation}
    \sigma_R^2 = \sum_{t=1}^T \sigma_{r_t}^2 + 2 \sum_{i<j} \mathrm{Cov}(r_i, r_j)
\end{equation}

In ORM, the covariance term is often positive due to the shared $R(\tau)$ and accumulates with $T$, causing the overall noise to grow rapidly with sequence length.
Specifically:
\begin{enumerate}
    \item \textbf{Uncorrelated Case:} If $\mathrm{Cov}(r_i, r_j) = 0$ for all $i \neq j$, then $\sigma_R^2 = T\sigma_r^2$ (linear growth).
    \item \textbf{Strongly Correlated Case:} If there exists $\mathrm{Cov}(r_i, r_j) \approx \sigma_r^2$ for all pairs, then $\sigma_R^2 = T^2 \sigma_r^2$ (quadratic growth).
\end{enumerate}

\subsubsection{Disentangling Credit Assignment from Variance.}
\label{sec:Variance Analysis}

A natural question arises: Is ORM's failure purely driven by this statistical confound, or fundamentally rooted in the structural credit assignment flaw?

To isolate and rule out the variance interference, we construct a variance-reduction baseline following the conventional PRM implementation in mathematical reasoning~\citep{lightman2023let}. Candidates are sampled at the paragraph level, but the advantage is computed over the full trajectory:

\begin{equation}
    A^{\mathrm{vanilla}}_{i,k} = 
    \frac{R(o_i) - \mu}{\sigma + \epsilon},
\end{equation}

where $R(o_i) = \sum_{k} r_{i,k}$ is the trajectory-level return and $\mu, \sigma$ are its mean and standard deviation across the $G$ candidates. We refer to this baseline as Vanilla PRM. By increasing the group size $G$ to match the number of segments per document, Vanilla PRM provides a strong test of the variance hypothesis within the ORM reward framework.

Table~\ref{tab:ablation_timing} shows that Vanilla PRM still underperforms SH-PRM despite its larger sampling budget, indicating that variance alone does not explain the gap. Variance reduction can suppress part of the statistical noise, but it cannot eliminate cross-segment reward coupling. As Equation~\eqref{eq:prm_grad} shows, the interference term $R_{-k}$ continues to couple each segment’s learning signal with rewards from other segments regardless of $G$. SH-PRM addresses this interference through supervision design rather than sampling alone.

% \textbf{[AAAI APPENDIX REF -- REVIEW BEFORE SUBMISSION]}

\section{Experimental Setup}
\label{sec:experimental_setup}
% We evaluate SH-PRM-GRPO on complex task of long-form report generation using a synthetic training set spanning 80 domains. Experiments are conducted on Qwen3 (8B/32B) and Llama-3-8B, and evaluated on DeepResearch Bench and WritingBench to assess both deep analytical capabilities and general writing proficiency.

\subsection{Datasets and Benchmarks}
\label{sec:data_benchmarks}
% \textbf{[MOVED FROM APPENDIX A -- REVIEW]}
% \textbf{[MOVED FROM APPENDIX A -- REVIEW]} 
% \textbf{[MOVED FROM APPENDIX A -- REVIEW]} 
\paragraph{Training Data.}

We construct a synthetic dataset of 6163 samples spanning 80 domains using a Wikipedia-based hierarchical pipeline that refines high-level domains into research queries, context summaries, and structured outlines with 3--6 chapters. This dataset serves as the prompt source for GRPO training. Detailed generation algorithms and statistics are provided in the supplementary material.

% \textbf{[AAAI APPENDIX REF -- REVIEW BEFORE SUBMISSION]}

% We construct a synthetic dataset of 1,378 samples spanning 80 domains using a Wikipedia-based hierarchical pipeline that refines high-level domains into research queries, context summaries, and structured outlines with 3--6 chapters. This dataset serves as the prompt source for GRPO training. Detailed generation algorithms and statistics are provided in Appendix~\ref{app:data_construction}.

\paragraph{Evaluation Benchmarks.}

  We evaluate the models on two complementary benchmarks for long-form generation. \textbf{DeepResearch Bench}~\cite{deep25} measures the ability to conduct deep research and synthesize information into structured, multi-chapter reports, while \textbf{WritingBench}~\cite{writing25} evaluates instruction-following under diverse long-form writing constraints. To ensure fair comparisons, we enforce a unified generation configuration across all baselines. 
  In particular, all models are evaluated under the same inference pipeline, with identical prompt structure, decoding settings, and pre-retrieved reference materials. For each DeepResearch Bench query, we freeze a shared set of reference documents and provide the same context to the base model, ORM, and SH-PRM variants, so performance differences reflect generation quality rather than retrieval variance.

  % \textbf{[AAAI APPENDIX REF -- REVIEW BEFORE SUBMISSION]}
  % We evaluate on two complementary benchmarks for long-form generation. \textbf{DeepResearch Bench} measures the ability to synthesize structured, multi-chapter reports, while \textbf{WritingBench} evaluates instruction-following and long-text generation under complex structural constraints. We use a unified generation configuration across all baselines; detailed evaluation prompts and parsing logic are provided in Appendix~\ref{app:racebench_eval}.

\subsection{Base Models}
  \label{sec:models}
  To evaluate both scalability and cross-architecture generalization, we conduct experiments on three base models: Qwen3-8B, Qwen3-32B, and Llama-3-8B. Implementation details, including hardware specifications and DeepSpeed configurations, are provided in the supplementary material.
% \textbf{[AAAI APPENDIX REF -- REVIEW BEFORE SUBMISSION]}
 \subsection{Baselines}
  \label{sec:baselines}
  We compare our method against four baselines spanning both inference-time and training-based settings, designed to isolate the effects of supervision granularity and rubric design under otherwise matched optimization settings:
  \begin{itemize}
      \item \textbf{Base Model:} The model evaluated with greedy decoding, serving as the performance lower bound.
      \item \textbf{Best-of-N (BoN):} A strong inference-time baseline that samples $N=8$ candidates and ranks them using the same rubric-based reward model. We set $N=8$ to match the GRPO group size.
      \item \textbf{ORM-GRPO:} The standard outcome-supervised RL baseline, trained under the same GRPO framework, training budget, and decoding protocol as SH-PRM, but receiving a single scalar reward only after the full document is generated. 
      \item \textbf{SH-PRM (No-Rubric):} A variant that retains process supervision but replaces the multi-dimensional rubric with a single holistic score, allowing us to evaluate the contribution of fine-grained reward decomposition.
  \end{itemize}
We omit comparisons with PPO or DPO variants, as our goal is to isolate the effect of supervision granularity under a shared optimizer rather than benchmark RL algorithms.

% -----------------------------------------------------------------
% 5.4 Reward Design and Training Configuration
% -----------------------------------------------------------------
\subsection{Training Configuration}
\label{sec:reward_hyperparams}

\paragraph{Gemini-based Rubric Reward.}
To implement the segment-level reward in Section~\ref{sec:multidimensional_rubric}, we employ a rubric evaluator~\cite{kim2023prometheus} with three scored dimensions. The three dimensions are each scored from 0 to 5, and the resulting segment-level reward follows Eq.~\ref{eq:segment_reward}. All compared methods in this study including Base, ORM-GRPO, and SH-PRM are trained and evaluated with the same Gemini-2.5-Flash rubric evaluator, ensuring that any systematic bias in the evaluator affects all conditions equally and does not confound the relative comparison of supervision granularity. We adopt Gemini following prior evidence of strong agreement with human evaluation~\cite{deep25}. To further verify that our conclusions are not specific to this evaluator family, we conduct a cross-evaluator check with Qwen-A30B. 
The relative ordering among all methods remains consistent (Base < ORM-GRPO < SH-PRM), with SH-PRM still achieving the strongest overall performance. 

% \textbf{[AAAI APPENDIX REF -- REVIEW BEFORE SUBMISSION]}
% \textbf{[AAAI APPENDIX REF -- REVIEW BEFORE SUBMISSION]}
\paragraph{Training Configuration.}
Following the SH-PRM-GRPO procedure, we use a group size of $G=8$ candidates per segment throughout our experiments. Additional hyperparameters are detailed in the supplementary material.
% \textbf{[AAAI APPENDIX REF -- REVIEW BEFORE SUBMISSION]}
\paragraph{Inference Strategy.}
To ensure a fair comparison of computational efficiency, all trained models are evaluated using \textbf{greedy decoding}~\cite{germann-2003-greedy}. We do not employ inference-time search strategies for the trained models, as our goal is to internalize reasoning capabilities into the model weights for efficient deployment.

\section{Results and Analysis}

\subsection{Main Results}
\label{sec:main result}
\begin{table*}[ht]
\vskip 0.02in
\begin{center}
\begin{small}
\setlength{\tabcolsep}{4.4pt}
\begin{tabular}{llcccccc}
\toprule
\multirow{2}{*}{\textbf{Model}} & \multirow{2}{*}{\textbf{Method}} & \multicolumn{5}{c}{\textbf{DeepResearch Bench}} & \textbf{WritingBench} \\
\cmidrule(lr){3-7} \cmidrule(lr){8-8}
& & \textbf{Overall} & \textbf{Comp.} & \textbf{Insight} & \textbf{Inst.} & \textbf{Read.} & \textbf{Score} \\
\midrule

\multirow{5}{*}{\textbf{Qwen3-8B}}
& Base              & 29.49 & 28.35 & 22.73 & 37.44 & 31.11 & 45.9 \\
& + Best-of-N       & 34.80 & 33.36 & 28.73 & 41.45 & 38.35 & 59.7 \\
& + ORM-GRPO        & 33.42 & 31.93 & 26.41 & 42.23 & 34.53 & 63.7 \\
& + SH-PRM (No-Rubric) & 34.55 & 33.05 & 28.10 & 42.35 & 37.20 & 60.7 \\
& + SH-PRM (Ours)      & \textbf{35.61} (+2.19) & \textbf{34.01} (+2.08) & \textbf{29.56} (+3.15) & \textbf{42.55} (+0.32) & \textbf{39.06} (+4.53) & \textbf{70.6} (+6.90) \\
\midrule

\multirow{5}{*}{\textbf{Qwen3-32B}}
& Base              & 35.85 & 33.67 & 29.50 & 43.70 & 38.24 & 66.5 \\
& + Best-of-N       & 35.92 & 33.85 & 29.65 & 43.65 & 38.50 & 67.9 \\
& + ORM-GRPO        & 36.03 & 34.10 & 29.78 & 43.63 & 38.90 & 69.5 \\
& + SH-PRM (No-Rubric) & 36.48 & 34.22 & 30.24 & 44.31 & 39.28 & 67.4 \\
& + SH-PRM (Ours)      & \textbf{38.84} (+2.81) & \textbf{36.91} (+2.81) & \textbf{33.15} (+3.37) & \textbf{45.79} (+2.16) & \textbf{41.82} (+2.92) & \textbf{72.7} (+3.20) \\
\midrule

\multirow{5}{*}{\textbf{Llama-3-8B}}
& Base              & 26.09 & 23.85 & 17.05 & 37.01 & 28.85 & 36.5 \\
& + Best-of-N       & 26.85 & 24.50 & 18.20 & 37.45 & 29.50 & 37.6 \\
& + ORM-GRPO        & 27.60 & 25.17 & 19.24 & 37.86 & 30.22 & 37.3 \\
& + SH-PRM (No-Rubric) & 29.55 & 27.20 & 22.10 & 38.25 & 33.10 & 45.6 \\
& + SH-PRM (Ours)      & \textbf{31.16} (+3.56) & \textbf{29.00} (+3.83) & \textbf{24.57} (+5.33) & \textbf{38.60} (+0.74) & \textbf{35.51} (+5.29) & \textbf{51.2} (+13.90) \\
\bottomrule

\end{tabular}
\end{small}
\end{center}
\caption{\textbf{Main results.} 
Comparison across model families on DeepResearch Bench and WritingBench. DeepResearch Bench reports \textbf{Overall}, \textbf{Comp.} (Completeness), \textbf{Insight}, \textbf{Inst.} (Instruction Following), and \textbf{Read.} (Readability), and all scores are reported on a 0--100 scale. \textbf{Bold} marks the best result within each family. \textit{Best-of-N} uses inference-time rejection sampling with $N=8$, and \textit{No-Rubric} denotes SH-PRM with a single holistic score. Parenthetical values indicate the absolute improvement of SH-PRM over the standard ORM-GRPO baseline.\label{tab:main_results}}
\end{table*}

\paragraph{Automated Evaluation Results.}
Table~\ref{tab:main_results} demonstrates that SH-PRM consistently achieves the best performance across all model families (Qwen3-8B/32B, Llama-3-8B) on both DeepResearch Bench and WritingBench. Compared to outcome-level RL and inference-time Best-of-N, SH-PRM yields substantial gains. For instance, on Qwen3-32B (DeepResearch Bench), SH-PRM reaches 38.84, outperforming ORM-GRPO (36.03) and Best-of-N (35.92). Similar trends hold for Qwen3-8B (35.61 vs. 33.42 and 34.80) and across all WritingBench evaluations.

% These comparisons isolate two key sources of improvement. First, the consistent gap between SH-PRM and ORM-GRPO directly validates the superiority of process-level supervision. Second, comparing SH-PRM with \textit{SH-PRM (No-Rubric)} highlights the necessity of multi-dimensional rewards: explicit rubrics boost the overall score from 36.48 to 38.84 on Qwen3-32B.

% Finally, SH-PRM significantly enhances the base models across all settings. The pronounced gains on both the highly capable Qwen3-32B and the smaller Llama-3-8B suggest that process supervision is universally beneficial: it guides strong models to exploit fine-grained feedback, while providing crucial structural scaffolding for weaker models.

% We also verify the same trend under a GSPO-based optimization variant on Qwen3-8B (Appendix~\ref{app:gspo_check}), suggesting that the gains are not tied to a specific group-based optimizer.
\paragraph{Manual Evaluation Results.}
We further conduct human evaluation on the full DeepResearch Bench test set. Multiple annotators rank the outputs of three models (Base, trained by ORM, and trained by SH-PRM) from best to worst. Detailed annotation guidelines and setup are provided in the appendix.

We aggregate all annotations using the mean rank:
\begin{equation}
\mathrm{MeanRank}(m) = \frac{1}{N \times R} \sum_{i=1}^N \sum_{r=1}^R \mathrm{rank}_{i,r}(m)
\label{eq:mean_rank}
\end{equation}
where $N$ is the number of evaluation samples and $R$ is the number of annotators. A smaller mean rank indicates better human preference.

As shown in Table \ref{tab:human_rank}, SH-PRM achieves the best mean rank of 1.84, outperforming both ORM and the base model. This result is consistent with the automatic evaluation and suggests that SH-PRM improves long-form content organization and overall response quality.
\begin{table}[t!]
\centering
\begin{small}
\begin{tabular}{lccc}
\toprule
\textbf{Model} & \textbf{Base} & \textbf{ORM-GRPO} & \textbf{SH-PRM (Ours)} \\
\midrule
Mean Rank & 2.72 & 2.41 & \textbf{1.84} \\
\bottomrule
\end{tabular}
\end{small}
\caption{Human evaluation mean rank results on DeepResearch Bench. Lower mean rank corresponds to better human preference.}
\label{tab:human_rank}
\end{table}

\subsection{Training Dynamics and Stability}

\begin{figure*}[t]
    \centering
    \includegraphics[width=0.98\textwidth]{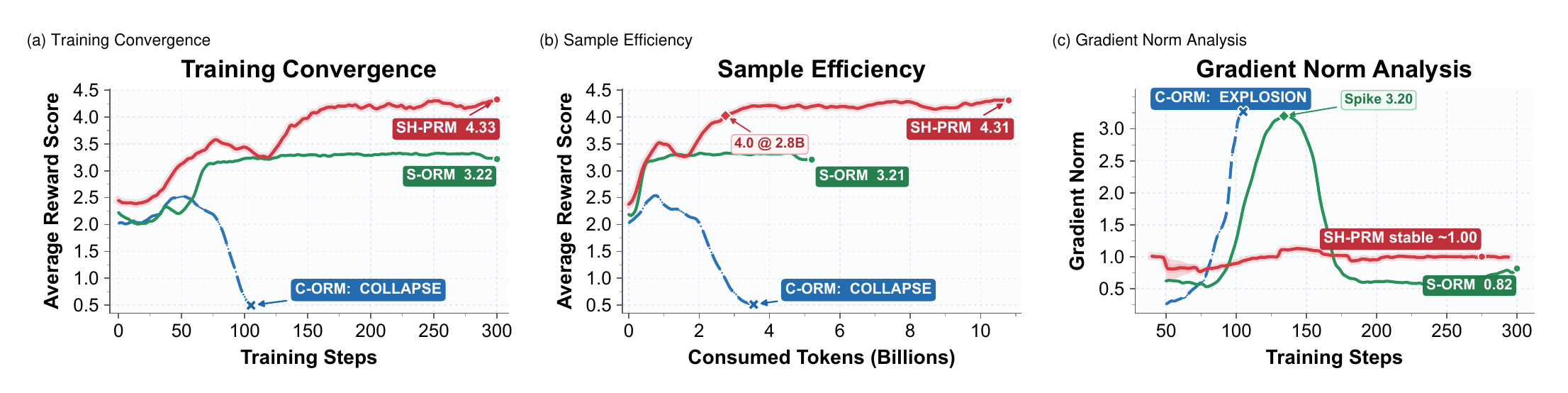}
    \caption{\textbf{Training dynamics and stability analysis.}
    \textbf{(a) \& (b):} Our SH-PRM method (Red) maintains a steady upward trajectory and superior sample efficiency, whereas ORM baselines either collapse or prematurely plateau. \textit{Note:} The y-axis represents a unified outcome quality score evaluated on a held-out validation set.
    \textbf{(c):} The microscopic cause of this difference. The \textit{Collapsed} ORM (Blue) exhibits gradient explosion, and the \textit{Stabilized} ORM (Green) shows a large gradient spike ($>3.0$). In contrast, our SH-PRM (Red) maintains a stable gradient norm ($\sim$1.0), indicating that segment-level rewards provide a much cleaner and more stable optimization signal.}
    \label{fig:training_dynamics_combined}
\end{figure*}

We analyze the training trajectories to understand how process supervision affects optimization stability and sample efficiency (Figure~\ref{fig:training_dynamics_combined}), and relate these macroscopic behaviors to the corresponding gradient dynamics in Figure~\ref{fig:training_dynamics_combined}(c).
\paragraph{Convergence and Efficiency.}
Figure~\ref{fig:training_dynamics_combined} shows the instability of training with sparse outcome signals. 
Without KL regularization, ORM (blue) collapses around step 80.  
Adding a KL penalty (Stabilized ORM, green) avoids collapse but the model quickly plateaus at a lower reward level ($\sim$3.3) after consuming about 2 billion tokens.
By contrast, SH-PRM (red) maintains a steady upward trajectory, converges to a higher reward level ($\sim$4.3), and continues improving as training proceeds.
Figure~\ref{fig:training_dynamics_combined}(c) further shows that SH-PRM (red) maintains a substantially more stable gradient norm ($\sim$1.0), whereas stabilized ORM (green) still exhibits a large gradient spike above 3.0 before plateauing. This behavior is consistent with the analysis in Section~\ref{sec:method_theoretical_analysis} that outcome-level interference grows with the number of segments, helping explain the smoother convergence of SH-PRM.

\paragraph{Training Efficiency.}
Although SH-PRM incurs slightly higher per-step cost than ORM, this overhead is offset by faster convergence, as shown in Table~\ref{tab:computational_efficiency}.
On Qwen3-8B, SH-PRM-GRPO takes 52.8 seconds per step versus 45.2 for ORM-GRPO, but reaches the target performance in 300 steps, whereas ORM remains unstable beyond 450 steps. 
This reduces total wall-clock training time from over 5.6 hours to 4.4 hours, indicating that faster convergence outweighs the modest increase in per-step cost.
At the token level, the extra per-step cost introduced by prefix-conditioned generation is largely offset by faster convergence, bringing the effective overall training token cost close to ORM in practice (see Appendix 6 for details).
Overall, these results show that SH-PRM is substantially more training-efficient under comparable computational budgets.

\subsection{Ablation Studies}
\label{sec:ablation}

To better understand the contribution of different design choices, we conduct ablation studies along three axes: rubric dimensions, reward timing, and segment granularity. Due to computational constraints, rubric ablations are conducted on Qwen3-32B, while reward-timing and granularity ablations are performed on Qwen3-8B.

\begin{table}[t]
\centering
\setlength{\tabcolsep}{3.4pt}
\begin{small}
\begin{tabular}{@{}lcccccc@{}}
\toprule
\textbf{Setting} & \textbf{Overall} & \textbf{Comp.} & \textbf{Insight} & \textbf{Inst.} & \textbf{Read.} & \textbf{WB} \\
\midrule
SH-PRM (Full) & \textbf{38.84} & \textbf{36.91} & \textbf{33.15} & \textbf{45.79} & \textbf{41.82} & \textbf{72.7} \\
w/o Rel. & 35.03 & 32.94 & 28.66 & 43.09 & 37.76 & 66.8 \\
w/o Coh. & 36.04 & 33.78 & 29.86 & 43.77 & 39.29 & 68.5 \\
w/o Adh. & 36.36 & 34.42 & 30.13 & 44.11 & 39.14 & 67.6 \\
\bottomrule
\end{tabular}
\end{small}
\caption{Ablation on rubric components using Qwen3-32B. Full Method refers to our standard SH-PRM with all three dimensions enabled.}
\label{tab:ablation_rubric}
\end{table}

\begin{table}[t]
\centering
\setlength{\tabcolsep}{3.2pt}
\begin{small}
\begin{tabular}{@{}lcccccc@{}}
\toprule
\textbf{Setting} & \textbf{Overall} & \textbf{Comp.} & \textbf{Insight} & \textbf{Inst.} & \textbf{Read.} & \textbf{WB} \\
\midrule
ORM & 33.42 & 31.93 & 26.41 & 42.23 & 34.53 & 63.7 \\
Vanilla PRM & 34.16 & 31.15 & 27.81 & 39.51 & 35.78 & 68.8 \\
SH-PRM & \textbf{35.61} & \textbf{34.01} & \textbf{29.56} & \textbf{42.55} & \textbf{39.06} & \textbf{70.6} \\
\bottomrule
\end{tabular}
\end{small}
\caption{Reward timing on Qwen3-8B: ORM uses document-level holistic scoring, Vanilla PRM aggregates segment-level scores at the end, and SH-PRM applies step-wise segment supervision.}
\label{tab:ablation_timing}
\end{table}

\begin{table}[t]
\centering
\setlength{\tabcolsep}{4.6pt}
\begin{small}
\begin{tabular}{@{}lcccccc@{}}
\toprule
\textbf{Setting} & \textbf{Overall} & \textbf{Comp.} & \textbf{Insight} & \textbf{Inst.} & \textbf{Read.} & \textbf{WB} \\
\midrule
$t=2$ seg. & 34.26 & 29.49 & 42.33 & 38.78 & 33.55 & 64.5 \\
$t=4$ seg. & 35.64 & 32.39 & \textbf{43.47} & \textbf{41.98} & 36.90 & 65.6 \\
$t=6$ seg. & \textbf{36.95} & \textbf{34.93} & 42.85 & 41.17 & \textbf{39.29} & \textbf{67.4} \\
$t=8$ seg. & 33.20 & 28.75 & 42.50 & 38.57 & 35.12 & 66.8 \\
$t=10$ seg. & 33.58 & 28.88 & 42.29 & 38.65 & 35.21 & 64.3 \\
\bottomrule
\end{tabular}
\end{small}
\caption{Impact of segment granularity ($t$) on Qwen3-8B. Scores are on a 0--100 scale.}
\label{tab:ablation_granularity}
\end{table}

\begin{table}[htbp]
\centering
\setlength{\tabcolsep}{10pt}
\begin{small}
\begin{tabular}{@{}lcr@{}}
\toprule
\textbf{Metric} & \textbf{ORM} & \textbf{SH-PRM} \\
\midrule
Time per step (seconds) & 45.2 & 52.8 \\
Steps to target & 450+ (unstable) & 300 \\
Total wall-clock time (hours) & 5.6+ & 4.4 \\
Overall training token cost & 1 & 1.16 \\
\bottomrule
\end{tabular}
\end{small}
\caption{Computational Efficiency Comparison (Qwen3-8B)}
\label{tab:computational_efficiency}
\end{table}

\paragraph{Impact of Rubric Dimensions.}
Table~\ref{tab:ablation_rubric} shows that all three rubric dimensions contribute to performance. Removing \textit{Relevance} causes the largest overall drop (38.84 $\rightarrow$ 35.03), suggesting that topical grounding is the strongest constraint in this setting. Removing \textit{Coherence} mainly hurts \textit{Readability} (41.82 $\rightarrow$ 39.29), while removing \textit{Adherence} mainly hurts \textit{Instruction Following} (45.79 $\rightarrow$ 44.11), indicating that the three dimensions play complementary roles.

\paragraph{Reward Timing.}
Table~\ref{tab:ablation_timing} compares outcome-level supervision (ORM), segment-level scoring aggregated at the end (Vanilla PRM), and step-wise process supervision (SH-PRM). The improvement from ORM (33.42) to Vanilla PRM (34.16) suggests that segment-level rubric signals already provide a more informative training signal than a single holistic reward. Compared with Vanilla PRM, SH-PRM yields stronger \textit{Instruction Following} (42.55 vs. 39.51), better overall performance (35.61 vs. 34.16), and higher WritingBench performance (70.6 vs. 68.8), suggesting that step-wise supervision provides stronger structural control.

\paragraph{Segment Granularity.}
Table~\ref{tab:ablation_granularity} shows a non-monotonic effect of segment granularity. Performance improves as the number of segments increases from $t=2$ to $t=6$, with the best \textit{Overall} score (36.95), \textit{Readability} (39.29), and WritingBench score (67.4) achieved at $t=6$, but drops again at $t=8$ and $t=10$. This suggests that overly coarse decomposition leaves rewards too sparse, while overly fine decomposition can hurt document-level coherence, making intermediate granularity the best balance.

\section{Conclusion}

This paper studies how supervision granularity shapes reinforcement learning for long-form generation. We propose SH-PRM, a short-horizon framework that reformulates long-form generation through semantic decomposition and trains the policy with process supervision over semantic segments. 
Across Qwen and Llama models, SH-PRM consistently improves generation quality and training efficiency under the same optimizer and comparable budgets. More broadly, our theoretical and empirical analyses clarify that the weakness of outcome-level supervision primarily stems from structural credit assignment rather than statistical variance alone. We hope this perspective is useful for future work on long-form generation and process supervision.

\newpage
\bibliography{aaai2027}
\appendix

\section{Data Construction Details}
\label{app:data_construction}

This section provides the detailed data-generation procedures, prompts, and validation artifacts corresponding to the training-data summary in the main paper. We retain the same top-down construction pipeline and focus here on the components omitted from the main text, including the explicit generation algorithm, prompt templates, and a representative case study.

\subsection{Generation Algorithm}

The overall procedure is summarized in Algorithm~\ref{alg:data_gen}. We employ robust parsing and validation steps to ensure data quality.

\subsection{Prompts Used}

\begin{algorithm}[t]
\caption{Hierarchical Synthetic Data Generation}
\label{alg:data_gen}
\begin{algorithmic}[1]
   \REQUIRE Target total samples $N$, Topics per domain $K$, Wikipedia Dump $\mathcal{W}$
   \ENSURE Dataset $\mathcal{D} = \{(q_i, c_i, o_i)\}_{i=1}^N$
   
   \STATE Initialize $\mathcal{D} \leftarrow \emptyset$
   \STATE $N_{\text{domains}} \leftarrow N / K$
   
   \STATE \textit{\# Step 0: Sample seed domains from real-world source}
   \STATE $\text{Domains} \leftarrow \text{Sample}(\mathcal{W}, N_{\text{domains}})$
   
   \FOR{\textbf{each} $\text{domain} \in \text{Domains}$}
       \STATE \textit{\# Step 1: Expand domain into specific research topics}
       \STATE $\text{Topics} \leftarrow \text{LLM}(\text{Prompt}_{\text{topic}}(\text{domain}))$
       
       \FOR{\textbf{each} $\text{topic} \in \text{Topics}$}
           \STATE \textit{\# Step 2: Enrich with Query and Summary}
           \STATE $(q, c) \leftarrow \text{LLM}(\text{Prompt}_{\text{enrich}}(\text{topic}))$
           
           \STATE \textit{\# Step 3: Generate Outline}
           \STATE $n_{\text{chapters}} \sim \text{Uniform}(3, 6)$
           \STATE $o \leftarrow \text{LLM}(\text{Prompt}_{\text{outline}}(q, c, n_{\text{chapters}}))$
           
           \IF{$o$ is valid}
               \STATE $\mathcal{D} \leftarrow \mathcal{D} \cup \{(q, c, o)\}$
           \ENDIF
       \ENDFOR
   \ENDFOR
   
   \RETURN $\mathcal{D}$
\end{algorithmic}
\end{algorithm}

We provide the specific prompts used in our pipeline below. Figure~\ref{fig:prompt_domain} shows the prompt for expanding the Wikipedia-sourced domains into specific research topics, while Figure~\ref{fig:prompt_outline} illustrates the prompt for generating the structured outline.

\newtcolorbox{promptbox}[1]{colback=gray!5!white,colframe=gray!75!black,title=#1,fonttitle=\bfseries}

% --- Prompt Box 1: Topic Generation ---
\begin{figure}[h]
\begin{promptbox}{Prompt for Step 1: Topic Generation}
\textbf{Context:} We have extracted the domain ``\{domain\}'' from Wikipedia. \\
\textbf{Role:} You are a research topic generator. \\
\textbf{Task:} Generate \{num\_topics\} specific, complex, and high-value research report topics within this domain. \\

\textit{Requirements:} \\
1. The topics should be suitable for a comprehensive, multi-chapter in-depth analysis (3 to 6 chapters). \\
2. Avoid simple questions. Use professional titles. \\
3. Ensure the topics are grounded in the real-world context of the domain. \\

\textit{Output Format:} Strictly a JSON list of objects containing ``title'' and ``category''.
\end{promptbox}
\caption{Prompt used to expand Wikipedia-sourced domains into specific research topics.}
\label{fig:prompt_domain}
\end{figure}

% --- Prompt Box 2: Outline Generation ---
\begin{figure}[h]
\begin{promptbox}{Prompt for Step 3: Structured Outline Generation}
You are an expert in structuring research reports.

\textbf{[User Query]} \\
\{query\}

\textbf{[Context / Summary]} \\
\textbf{Title:} \{title\} \\
\textbf{Summary:} \{summary\}

\textbf{Task:} Design a \{num\_chapters\}-chapter outline for this report.

\textit{Requirements:} \\
1. Create exactly \{num\_chapters\} top-level chapter titles. \\
2. Format: \\
\hspace*{0.5cm} 1. [Chapter Title] \\
\hspace*{0.5cm} 2. [Chapter Title] \\
\hspace*{0.5cm} ... \\
\hspace*{0.5cm} \{num\_chapters\}. [Chapter Title] \\
3. Logical Flow: Ensure a coherent progression from introduction to conclusion. \\
4. Be specific and informative.

\textit{Output:} Only the \{num\_chapters\} lines of the outline. No other text.
\end{promptbox}
\caption{Prompt used to generate the structured outline, serving as the ground truth or reference for GRPO training.}
\label{fig:prompt_outline}
\end{figure}

\subsection{Case Study}
\label{app:case_study}

To demonstrate the quality and complexity of our synthetic dataset, Table~\ref{tab:case_study} presents a representative example generated by our pipeline. This example illustrates how a high-level domain sampled from Wikipedia (``Quantum Computing'') is progressively refined into a specific, technical research topic, enriched with a detailed user query and context, and finally structured into a coherent multi-chapter outline.

\begin{table*}[h]
\caption{A real example from our synthetic dataset. The pipeline starts from a Wikipedia-sourced domain and generates a structured research outline suitable for a complex technical report.}
\label{tab:case_study}
\begin{center}
\begin{small}
\begin{tabular}{p{0.18\textwidth} p{0.78\textwidth}}
\toprule
\textbf{Stage} & \textbf{Content} \\
\midrule
\textbf{1. Seed Domain} & \textit{Quantum Computing} (Sampled from Wikipedia) \\
\midrule
\textbf{2. Generated Topic} & \textbf{Title:} Architectural Design and Protocol Development for a Global Quantum Internet: Challenges and Prospects for Secure Communication \\
& \textbf{Category:} Quantum Computing \\
\midrule
\textbf{3. Enriched Input} & \textbf{User Query:} Write a comprehensive report on `Architectural Design and Protocol Development for a Global Quantum Internet: Challenges and Prospects for Secure Communication.' Your report should cover fundamental architectural paradigms, including network topologies and hardware integration, as well as key protocol layers necessary for secure quantum communication (e.g., Quantum Key Distribution, entanglement distribution, routing). Analyze the significant technical and logistical challenges, such as qubit coherence, quantum repeater development, scalability, and standardization. Finally, discuss the transformative prospects for secure communication, outlining potential applications, security enhancements, and broader societal implications of a fully realized global quantum internet. \\
\vspace{0.1cm}
& \textbf{Context (Summary):} Architectural Design and Protocol Development for a Global Quantum Internet focuses on creating the foundational infrastructure and communication rules for a worldwide quantum network. This ambitious endeavor aims to leverage quantum phenomena like entanglement and superposition to enable ultra-secure communication, primarily through Quantum Key Distribution (QKD), and potentially distributed quantum computing. Key challenges include maintaining qubit coherence over long distances, developing robust quantum repeaters, and standardizing protocols across diverse quantum hardware. Despite these hurdles, the prospect of unbreakable cryptography, enhanced sensing, and novel distributed quantum applications drives intense global research and development efforts, promising a revolution in secure information exchange. \\
\midrule
\textbf{4. Target Outline} & 
1. Foundations and Vision: The Imperative for a Global Quantum Internet in Secure Communication \\
& 2. Architectural Paradigms and Protocol Development for Quantum Network Operation \\
& 3. Technical and Logistical Hurdles: Challenges in Building a Global Quantum Internet \\
& 4. Transformative Prospects and Societal Impact of Advanced Quantum Communication \\
\bottomrule
\end{tabular}
\end{small}
\end{center}
\end{table*}
\subsection{Data Statistics and Availability}
\label{app:stats}

\textbf{Quality Control.} To ensure the high quality of the training data, we implemented a rigorous filtering mechanism during the generation process. Specifically:
\begin{itemize}
    \item \textbf{Format Validation:} We discarded any samples where the LLM failed to produce valid JSON or where the outline did not match the requested number of chapters.
    \item \textbf{Length Constraints:} We filtered out summaries shorter than 50 words to ensure sufficient context for the model.
    \item \textbf{Logical Consistency:} We manually inspected a random subset of 50 samples to verify that the generated outlines were logically consistent with the user queries.
\end{itemize}

\textbf{Dataset Statistics.} The final dataset consists of $N=1378$ samples spanning 80 distinct domains. Each report contains 3--6 sections (4.2 on average), which yields 6,163 segment-level training instances under our training formulation. 
The distribution of outline lengths is as follows: 3 chapters (25\%), 4 chapters (35\%), 5 chapters (30\%), and 6 chapters (10\%). This diversity in length encourages the model to learn structural planning across varying levels of complexity. We observed that this dataset scale enables effective training convergence, and thus did not pursue further expansion. Nevertheless, our approach is designed to facilitate straightforward augmentation of the dataset should it be required.

\textbf{Availability.} For reproducibility, we provide an anonymous code repository.\footnote{\url{https://anonymous.4open.science/r/Prm-2159/}} Additional dataset artifacts are included in the supplementary material.

\section{Evaluation Details}
\label{app:racebench_eval}

To ensure a rigorous and fair comparison across all models (including the Base Model, ORM, and our SH-PRM), we implemented a unified evaluation pipeline using the DeepResearch Bench dataset. This section details the inference infrastructure, the unified prompt templates, and the input construction logic.

\subsection{Inference Pipeline and Configuration}
\label{app:inference_config}

All evaluations were conducted using the \texttt{vLLM} library to ensure high-throughput and deterministic generation. The specific inference configurations are as follows:

\begin{itemize}
    \item \textbf{Context Window:} We set the maximum context length to \textbf{32768 tokens} to accommodate the extensive search results and reference documents required for long-form reporting.
    \item \textbf{Generation Limits:} The maximum output length was restricted to \textbf{16484 tokens} to allow for sufficient depth in the generated reports.
    \item \textbf{Sampling Parameters:} To balance creativity and factual adherence, we used a temperature of $T=0.6$ and Top-P sampling with $p=0.95$.
\end{itemize}

\paragraph{Dynamic Truncation Strategy.}
Given the potentially massive volume of retrieved reference materials, we implemented a dynamic truncation strategy to prevent context overflow. The system prioritizes the retention of the System Prompt, User Query, Outline, and Task Descriptions. If the total token count exceeds the limit ($32,768 - 16484 - 500$ buffer), we truncate the \textit{References} section from the tail end while preserving the most relevant top-ranked chunks.

\subsection{Unified System Prompts}
\label{app:system_prompts}

To ensure a rigorous and fair comparison across all models (including the Base Model, ORM, and our SH-PRM), we implemented a unified evaluation pipeline using the \textbf{DeepResearch Bench} dataset.

To eliminate prompt engineering as a confounding variable, we employed a single, unified system prompt for all baselines and our method.

We provide the specific system prompt used to guide the models in generating the final long-form reports in Figure~\ref{fig:prompt_report_gen}. This prompt enforces strict constraints on source grounding and formatting to ensure consistency across all evaluations.

\begin{figure}[h]
    \centering
    \begin{promptbox}{Unified System Prompt: Report Generation}
        \textbf{Role:} You are a professional Report Generation Agent. \\
        \textbf{Task:} Write a well-organized, in-depth professional report based strictly on the provided Outline and Reference Materials. \\
        \textit{Core Instructions:} \\
        1. \textbf{Source Grounding:} You must rely \textbf{exclusively} on the provided reference documents to support your arguments. Do not hallucinate external information not present in the context. \\
        2. \textbf{Formatting:} Output the final report in Markdown, wrapped in \texttt{<final\_answer>} tags. Use \texttt{\#} for the main title and \texttt{\#\# 1.}, \texttt{\#\#\# 1.1.} for section headers. \\
        3. \textbf{Structure:} Begin with a concise abstract summarizing the core conclusions. Ensure the report is comprehensive and logically structured according to the provided outline.
    \end{promptbox}
    \caption{The unified system prompt used for report generation across all models (Base, ORM, and SH-PRM) to ensure evaluation fairness.}
    \label{fig:prompt_report_gen}
\end{figure}

\subsection{Context Consistency}
\label{app:context_consistency}

Since the models operate in an offline setting without internet access, ensuring consistent information availability is critical. For every test query in DeepResearch Bench, we pre-retrieved a fixed set of reference documents using a commercial search engine API. These references were frozen and provided identically to all models (Base, ORM, and SH-PRM) within the input context window. This guarantees that any performance difference is attributable solely to the model's reasoning and writing capabilities, rather than differences in information retrieval.

\subsection{Human Evaluation Details}
\subsubsection{Human Evaluation Setup}
For each query in DeepResearch Bench, we collected one response from each of the three compared systems: the base model, ORM-GRPO, and SH-PRM. This yields three candidate long-form answers per query. To eliminate position bias, we randomly shuffled the order of the three responses before annotation.

We then asked human annotators to rank the three responses by overall quality in a blind evaluation setting, producing a strict preference order such as 
2>1>3, where 1, 2, and 3 refer only to the displayed positions of the responses after shuffling. Annotators were instructed to provide a complete ranking for every query.

Our human evaluation involved three annotators with substantial experience in text assessment. All annotators were familiar with evaluating long-form writing and had prior experience with qualitative text evaluation tasks.

Across the full DeepResearch Bench test set of 100 queries, each query was evaluated by all three annotators, resulting in 300 complete rankings and 900 model-level rank assignments.
\subsubsection{Human Evaluation Results}
We report both \textbf{mean rank} (lower is better) and \textbf{pairwise win rate}. The mean-rank results in the main text show that SH-PRM achieves the best overall human preference among the three systems. Pairwise win rates further support this result:
\begin{itemize}
    \item SH-PRM beats ORM-GRPO in 56\% of comparisons;
    \item SH-PRM beats the Base model in 60\% of comparisons;
    \item ORM-GRPO beats the Base model in 56\% of comparisons.
\end{itemize}

These human judgments are consistent with the automatic evaluation results and further support the advantage of SH-PRM.
\section{Hyperparameter Settings}
\label{app:hyperparams}

In this section, we detail the training configurations and computational environment used for our experiments. All models were trained using the \texttt{trl} library with the GRPO algorithm.

\subsection{Training Hyperparameters}
We use a consistent set of hyperparameters across our experiments. The specific values are listed in Table~\ref{tab:hyperparams}. We utilize the AdamW optimizer with a cosine learning rate scheduler and a warmup ratio of 0.03.

\begin{table}[h]
    \centering
    \caption{Hyperparameters used for GRPO training.}
    \label{tab:hyperparams}
    \begin{tabular}{lc}
        \toprule
        \textbf{Hyperparameter} & \textbf{Value} \\
        \midrule
        \multicolumn{2}{c}{\textit{Optimization}} \\
        Learning Rate & $1 \times 10^{-6}$ \\
        LR Scheduler & Cosine \\
        Warmup Ratio & 0.03 \\
        Max Training Steps & 500 \\
        Optimizer & AdamW \\
        Precision & bfloat16 (BF16) \\
        \midrule
        \multicolumn{2}{c}{\textit{GRPO Specifics}} \\
        Group Size ($G$) & 8 \\
        Num Generations per Prompt & 8 \\
        \midrule
        \multicolumn{2}{c}{\textit{Batch Size \& Gradient}} c\\
        Per-Device Train Batch Size & 1 \\
        Gradient Accumulation Steps & 8 \\
        Gradient Checkpointing & True \\
        \bottomrule
    \end{tabular}
\end{table}

\subsection{Computational Environment and DeepSpeed Configuration}

All experiments were conducted on a single node equipped with 8 NVIDIA GPUs. To efficiently handle the memory requirements, particularly for the larger Qwen3-32B model, we employed DeepSpeed ZeRO Stage 3 (Zero Redundancy Optimizer)~\cite{rajbhandari2020zero}.

The specific DeepSpeed ZeRO-3 configuration used in our experiments is summarized in Table~\ref{tab:deepspeed_config}.

\begin{table}[htbp]
\centering
\caption{DeepSpeed ZeRO-3 Configuration Parameters}
\label{tab:deepspeed_config}
\begin{tabular}{@{}lc@{}} % @{} 去除表格左右两端的额外空白
\toprule
\textbf{Configuration Parameter} & \textbf{Value} \\
\midrule
\textbf{ZeRO Optimization (Stage 3)} & \\
\quad Offload Optimizer Device & CPU \\
\quad Offload Optimizer Pin Memory & True \\
\quad Offload Parameter Device & CPU \\
\quad Offload Parameter Pin Memory & True \\
\quad Overlap Communication & True \\
\quad Contiguous Gradients & True \\
\quad Sub-group Size & $1 \times 10^9$ \\
\midrule
\textbf{Precision} & \\
\quad BF16 Enabled & True \\
\bottomrule
\end{tabular}
\end{table}

\section{Reward Model Details}
\label{app:reward_prompts}

To ensure reproducible reward signals, we utilized a rubric-based prompting strategy with \textbf{Gemini-2.5-Flash} as the evaluator. We designed distinct prompts for our proposed SH-PRM and the ORM baseline.

\subsection{SH-PRM: Multi-dimensional Rubric Prompt}
For our proposed method, we evaluate each generated reasoning step (or section) based on three specific dimensions: \textit{Adherence}, \textit{Coherence}, and \textit{Relevance}. The scores are averaged to produce the final step-level reward.

\begin{figure}[h]
    \centering
    \begin{promptbox}{System Prompt: PRM Evaluator}
        
        \textbf{Role:} You are a strict scoring machine. Please output ONLY standard JSON format without any additional explanation. \\
        
        \textbf{User Prompt:} \\
        You are a professional writing evaluation expert. Please evaluate the quality of the following generated paragraph. \\
        
        \textbf{[Input Context]:} \\
        \{prompt\} \\
        
        \textbf{[Generated Paragraph]:} \\
        \{completion\} \\
        
        Please rate the content on the following three dimensions (0-5 scale, one decimal place supported): \\
        
        1. \textbf{Adherence Score:} Does the content strictly revolve around the section title? \\
        2. \textbf{Coherence Score:} Is the logic smooth? Is the transition from the context natural? \\
        3. \textbf{Relevance Score:} Does it answer the core requirements of the Query? \\
        
        \textit{Output Format:} \\
        \{ \\
        \hspace*{1em} "adherence\_score": 2.5, \\
        \hspace*{1em} "coherence\_score": 3.0, \\
        \hspace*{1em} "relevance\_score": 2.0 \\
        \}
        
    \end{promptbox}
    \caption{The multi-dimensional rubric prompt used for the PRM reward signal.}
    \label{fig:prm_prompt}
\end{figure}

\subsection{ORM: Outcome Quality Prompt}
For the ORM baseline, we employ a holistic evaluation prompt that focuses on the final usability and fluency of the text, rather than the step-by-step reasoning logic.

\begin{figure}[h]
    \centering
    \begin{promptbox}{System Prompt: ORM Evaluator}
        
        \textbf{Role:} You are a strict evaluator. Output only JSON. \\
        
        \textbf{User Prompt:} \\
        You are an expert evaluator assessing the quality of a generated report section. \\
        
        \textbf{[Input Context]:} \\
        ...\{truncated\_prompt\} \\
        
        \textbf{[Generated Section]:} \\
        \{completion\} \\
        
        Please evaluate the \textbf{Final Outcome Quality} of the generated section based on the context. Focus on: \\
        1. \textbf{Usability:} Is this text ready to be used in a final report? \\
        2. \textbf{Relevance:} Does it directly address the immediate instruction? \\
        3. \textbf{Fluency:} Is it well-written and professional? \\
        
        Do NOT evaluate the reasoning steps or the thought process. Only evaluate the final output text. \\
        
        Rate the text on a scale of 0.0 to 5.0. \\
        
        \textit{Output Format:} \\
        \{ \\
        \hspace*{1em} "overall\_score": <float> \\
        \}
        
    \end{promptbox}
    \caption{The holistic evaluation prompt used for the ORM baseline.}
    \label{fig:orm_prompt}
\end{figure}

\subsection{Evaluator Choice and Cross-Evaluator Check}
\label{sec:Evaluator Choice and Cross-Evaluator Check}

We use Gemini-2.5-Flash as the primary reward evaluator in the main experiments because prior work suggests that Gemini-based judgments are relatively well aligned with human preferences in open-ended generation settings. This makes it a practical choice for rubric-based segment evaluation in our setting.

To examine whether the main findings are overly dependent on a single evaluator, we additionally conduct a limited cross-evaluator comparison using Qwen-A30B on the Qwen3-8B setting. As shown in Table~\ref{tab:qwen_a30b_eval}, the relative ordering among the base model, ORM-GRPO, and SH-PRM remains broadly consistent with the Gemini-based evaluation, with SH-PRM still achieving the strongest overall performance. We therefore view this comparison as a robustness check suggesting that the main empirical trend is not purely an artifact of one specific API, while noting that it is limited in scope and should not be interpreted as a definitive evaluator study.

\begin{table*}[h]
\caption{Limited cross-evaluator comparison using Qwen-A30B on Qwen3-8B (DeepResearch Bench). Scores are reported on a 0--100 scale.}
\label{tab:qwen_a30b_eval}
\small
\begin{tabular*}{\textwidth}{@{\extracolsep{\fill}}lccccc}
\toprule
\textbf{Method} & \textbf{Overall} & \textbf{Comp.} & \textbf{Insight} & \textbf{Inst.} & \textbf{Read.} \\
\midrule
Qwen3-8B (Base) & 33.95 & 34.90 & 31.47 & 38.61 & 36.84 \\
Qwen3-8B-ORM-GRPO & 35.01 & 34.92 & 31.25 & 38.17 & 37.40 \\
Qwen3-8B-SH-PRM-GRPO (Ours) & \textbf{36.20} & \textbf{35.21} & \textbf{31.46} & \textbf{38.79} & \textbf{37.66} \\
\bottomrule
\end{tabular*}
\end{table*}

\section{Training Objective and Implementation Details}
\label{app:training_details}

\begin{figure*}[ht]
    \vspace{0.5cm}
    \centering
    \includegraphics[width=0.8\textwidth]{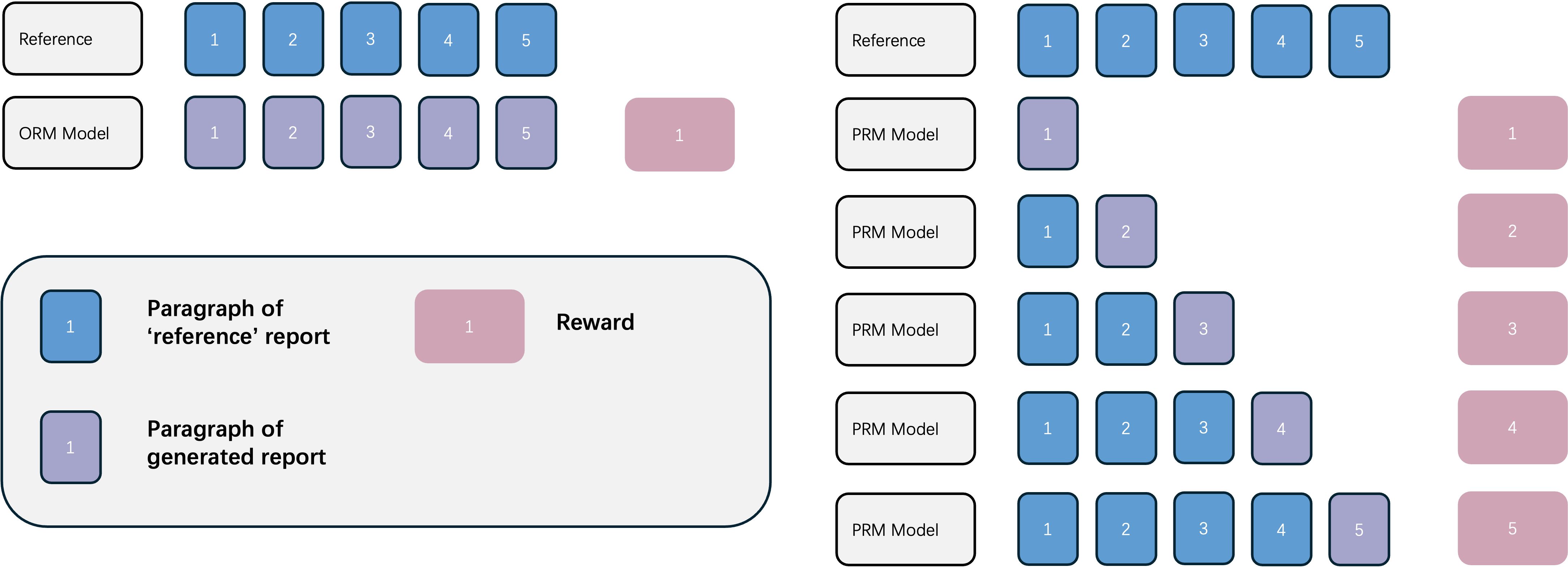}
    \caption{\textbf{Comparison of Reward Mechanisms: Outcome Reward Model (ORM) vs. Short-Horizon Process Reward Model (SH-PRM).} The \textbf{Left Panel} illustrates the ORM. It compares the entire generated report (paragraphs 1-5) against the reference to produce a single, sparse reward signal at the end. This makes it difficult to assign credit or blame to specific intermediate steps. The \textbf{Right Panel} depicts our SH-PRM. It provides dense, segment-level rewards by comparing each generated paragraph (e.g., paragraph 1) with its corresponding reference. This fine-grained feedback allows for precise credit assignment, guiding the model's learning process more effectively at each step of long-form generation.}
    \label{fig:reward_mechanism}
\end{figure*}

As illustrated in Figure~\ref{fig:reward_mechanism}, our SH-PRM fundamentally differs from the ORM by providing dense, segment-level feedback. This section provides comprehensive details necessary for reproducing our SH-PRM-GRPO framework, structured into data formulation, the formal training objective, and system infrastructure.

\subsection{Data Formulation and Reward Mechanism}

\paragraph{Context Formulation.} The training data is constructed from a curated dataset of 1,378 research reports spanning 80 domains, with each report decomposed into an average of 4.2 sections, yielding 6,163 segment-level training instances in total. To enable the model to learn section generation conditioned on any position within a document, each training sample corresponds to exactly one section. Specifically, the input context $s_k$ for the $k$-th section is formulated by concatenating: (1) the user query describing the research topic; (2) the complete outline specifying all $L$ section titles; (3) the target section title $c_k$ for the current segment; and (4) the ground-truth content of all previous sections $y_{<k} = (y_1, \dots, y_{k-1})$.

\paragraph{Reward Model Implementation.} We implement the process reward model using the Gemini-2.5-Flash API. For each candidate segment $y_k^{(i)}$, the API prompt is constructed using system instructions defining the evaluator role, alongside the input context ($q, c_k, y_{<k}$) and the candidate content $y_k^{(i)}$. The model evaluates the segment based on a detailed rubric covering Adherence, Coherence, and Relevance, outputting a JSON object with three scores on a 0-5 scale. To ensure strict reproducibility, we set the sampling temperature to 0 with fixed random seeds.

\subsection{System Infrastructure and Hyperparameters}

\paragraph{Training Infrastructure.} We implement SH-PRM-GRPO using the TRL library integrated with DeepSpeed ZeRO-3, distributed across eight A100 GPUs. To optimize memory usage and throughput, candidate generation ($G=8$) is split into two micro-batches of 4. Reward computation is parallelized asynchronously using a dedicated process pool of 32 workers to minimize API waiting time. Model checkpoints are saved every 100 steps for evaluation. With this configuration, training the Qwen3-8B model for 500 steps takes approximately 4 hours.

\paragraph{Hyperparameters.} The key hyperparameters used in our SH-PRM-GRPO training are summarized in Table~\ref{tab:hyperparams}. We utilize an effective batch size of 8, achieved through a per-device batch size of 1 and 8 gradient accumulation steps.

\begin{table}[htbp]
\centering
\caption{SH-PRM-GRPO Hyperparameters}
\label{tab:hyperparams}
\begin{tabular}{lc}
\toprule
\textbf{Hyperparameter} & \textbf{Value} \\
\midrule
Learning rate & $1 \times 10^{-6}$ \\
Learning rate scheduler & Cosine \\
Batch size per device & 1 \\
Gradient accumulation steps & 8 \\
Total batch size & 8 \\
Group size $G$ & 8 \\
Number of training steps & 500 \\
KL penalty coefficient $\beta$ & 0.01 \\
Clipping parameter $\epsilon$ & 0.2 \\
Maximum sequence length & 4096 \\
Precision & BF16 \\
\bottomrule
\end{tabular}
\end{table}

\section{Computational Cost Analysis}
\label{app:cost_analysis}
This appendix provides a detailed analysis of the computational costs associated with SH-PRM compared to ORM under GRPO. We systematically evaluate the token budget, training efficiency, and FLOPs to demonstrate the practical viability and advantages of the SH-PRM-GRPO approach.

\subsection{Token Budget Analysis}
Using the QwenTokenizer, we obtain the following averages per sample: the user query contains $\sim$550 tokens, section titles $\sim$97 tokens, and each paragraph $\sim$1,480 tokens. Each query corresponds to 4.33 paragraphs on average.

\paragraph{Per-step Token Cost.}
% In typical settings, an ORM sample corresponding to a full report can be split by paragraphs into multiple PRM samples. As a first-order approximation, the annotation-token ratio between ORM and PRM is therefore determined by the number of paragraphs, yielding $\sim$4.33:1.

In a special case without references, we estimate the per-step token cost ratio as:
\begin{equation}
\begin{aligned}
    \text{ORM} : \text{SH-PRM} &= \frac{L_q + L_t \cdot N_{\text{para}}}{\big[L_q + (L_t + L_p) \cdot N_{\text{para}}\big] / 4.33} \\[1ex]
    &\approx \frac{970}{1700} \approx 1 : 1.75
\end{aligned}
\end{equation}
where $L_q$, $L_t$, and $L_p$ denote the token lengths of the query, title, and paragraph, respectively. When reference documents are included, the additional  tokens appear in both the numerator and denominator, further reducing the ratio and making it closer to 1:1.

\paragraph{Quantitative Cost Estimate.}
Assuming a SH-PRM training set of size $N=1,378$ samples and a maximum per-report generation length of $L_{\max}=16,848$ tokens, using cost-effective models (e.g., Gemini-2.5-Flash or GPT-4o-mini at $\sim$\$0.60 per 1M tokens), the total generation cost for SH-PRM data construction is below \$14.00. 

\paragraph{Overall Training Token Cost.}
Since SH-PRM reaches the target performance in 300 steps while ORM remains unstable beyond 450 steps, the effective overall training token cost becomes $1.75 \times 300 : 1 \times 450 \approx 1.16 : 1$ (SH-PRM:ORM), showing that the total token budget remains close in practice despite the higher per-step cost of SH-PRM.

\paragraph{Data Efficiency.}
A naive sample-count comparison can be misleading, because one report-level sample is decomposed into multiple segment-level instances under SH-PRM, making it appear to use more samples. In fact, 300 SH-PRM segment-level steps correspond to only $300/4.33 \approx 69$ effective report-level samples, whereas ORM requires more than 450 full-report steps and still remains unstable. This indicates substantially higher data efficiency for SH-PRM.

\paragraph{Reward Evaluation.}
In principle, ORM evaluation would aggregate evaluations across all PRM steps. In practice, the number of criteria is stable (5--8), so the evaluation workload per step is approximately the same for ORM and SH-PRM.

\subsection{Training Efficiency Comparison}

Table~\ref{tab:computational_efficiency} summarizes the computational efficiency comparison between ORM-GRPO and SH-PRM-GRPO on Qwen3-8B. 

\begin{table}[htbp]
\centering
\caption{Computational Efficiency Comparison (Qwen3-8B)}
\label{tab:computational_efficiency}
\begin{tabular}{lcc}
\toprule
\textbf{Metric} & \textbf{ORM} & \textbf{SH-PRM} \\
\midrule
Time per step (seconds) & 45.2 & 52.8 \\
Steps to target & 450+ (unstable) & 300 \\
Total wall-clock time (hours) & 5.6+ & 4.4 \\
\bottomrule
\end{tabular}
\end{table}

SH-PRM requires fewer training steps to reach the target performance, leading to higher practical training efficiency. This shows that the modest increase in per-step cost is more than offset by faster convergence.

\subsection{FLOPs Analysis}

We provide an approximate FLOPs breakdown for forward pass, backward pass, and reward evaluation:

\begin{itemize}
    \item \textbf{Forward pass (generation):} ORM generates full text ($\sim$6,400 tokens), SH-PRM generates one paragraph ($\sim$1,480 tokens). Ratio: $\sim$4.3:1.
    \item \textbf{Backward pass:} Both use the same GRPO objective. SH-PRM computes gradients for one segment; ORM for full trajectory.
    \item \textbf{Reward evaluation:} SH-PRM evaluates 3 dimensions per paragraph; ORM evaluates once per trajectory. Total evaluation tokens: SH-PRM $\sim$1.5$\times$ ORM due to per-paragraph context overhead.
\end{itemize}

Overall, under a fixed computational budget B, SH-PRM enables substantially more localized updates than ORM, which is consistent with the credit-assignment analysis in Theorem 3.3 of the main text.

\section{Qualitative Case Study}
\label{app:qualitative_case}

To provide a tangible comparison between our method and the baseline, we present a side-by-side analysis of reports generated by \textbf{Qwen3-32B-SH-PRM (Ours)} and \textbf{Qwen3-32B-ORM (Baseline)} for the query: \textit{``Investigate the investment and lending relationships between domestic financial institutions and their connection to systemic risk.''}
We highlight two decisive advantages of our method: \textbf{Logical Structuring} and \textbf{Formatting Stability}.
\subsection{Advantage 1: Logical Structuring and Depth}
As shown in Figure~\ref{fig:case_structure}, the SH-PRM-trained model demonstrates superior capability in organizing complex concepts. 
In \textbf{Section 2.3 (Risk Contagion)}, the SH-PRM model explicitly categorizes the contagion paths into \textit{``Direct Contagion''} and \textit{``Indirect Contagion,''} using bold headers to guide the reader. This structural decomposition reflects a deeper understanding of the mechanism. In contrast, the ORM baseline produces a dense, unstructured paragraph (a ``wall of text'') that mixes definitions with examples, increasing the cognitive load for the reader.

\begin{figure*}[h]
    \centering
    \begin{tabular}{p{0.48\textwidth}|p{0.48\textwidth}}
        \toprule
        \multicolumn{1}{c|}{\textbf{Ours (SH-PRM-32B)}} & \multicolumn{1}{c}{\textbf{Baseline (ORM-32B)}} \\
        \midrule
        \textbf{Section 2.3 Risk Contagion Paths} & \textbf{Section 2.3 Risk Contagion Paths} \\
        Risk spreads among financial institutions through two main paths: & Risk contagion is mainly realized through channels such as the interbank network and the bond market. \\
        1. \textbf{Direct Contagion}: Direct transmission through credit default in the interbank lending market... & The capital lending and credit support in the interbank network are direct contagion media, while the bond market indirectly transmits the credit risk of large banks... \\
        2. \textbf{Indirect Contagion}: Realized through the bond market pattern of ``Large banks lend, small banks buy bonds''... & For example, in the Baoshang Bank event, due to the illegal occupation of funds by major shareholders... \\
        \vspace{0.1cm}
        \textit{\textbf{Analysis:} The SH-PRM model explicitly structures the argument into ``Direct'' and ``Indirect'' categories using bold formatting, making the mechanism clear and easy to follow.} & \vspace{0.1cm}
        \textit{\textbf{Analysis:} The ORM model buries the logic within a dense paragraph. It lists examples but lacks a clear conceptual framework for the contagion mechanisms.} \\
        \bottomrule
    \end{tabular}
    \caption{\textbf{Comparison of Logical Structuring.} Our model (Left) employs structural formatting to clarify complex concepts, whereas the baseline (Right) tends to generate unstructured text blocks.}
    \label{fig:case_structure}
\end{figure*}

\subsection{Advantage 2: Formatting Stability and Robustness}
Figure~\ref{fig:case_formatting} illustrates the robustness of our method against formatting hallucinations.
In \textbf{Section 4.1 (Risk Modeling)}, the SH-PRM model correctly utilizes bullet points to list different modeling approaches, ensuring high readability. Conversely, the ORM baseline exhibits signs of \textbf{reward hacking} and instability: it fails to use list structures and, in raw outputs, occasionally leaks special tokens (e.g., \texttt{<example>}, \texttt{<final\_answer>}) or enters repetition loops. This suggests that sparse outcome rewards are insufficient to constrain the model's generation format effectively.

\begin{figure*}[h]
    \centering
    \begin{tabular}{p{0.48\textwidth}|p{0.48\textwidth}}
        \toprule
        \multicolumn{1}{c|}{\textbf{Ours (SH-PRM-32B)}} & \multicolumn{1}{c}{\textbf{Baseline (ORM-32B)}} \\
        \midrule
        \textbf{Section 4.1 Mainstream Risk Modeling} & \textbf{Section 4.1 Mainstream Risk Modeling} \\
        Mainstream models include Complex Networks, CoVaR, and MES: & Complex network models, CoVaR (Conditional Value at Risk), and MES (Marginal Expected Shortfall) are currently mainstream risk modeling methods. \\
        - \textbf{Complex Network Model}: Quantifies risk contagion paths... & Complex network models identify key nodes... CoVaR is used to measure the average loss... while MES measures potential loss size... \\
        - \textbf{CoVaR}: Measures the conditional risk level... & \\
        - \textbf{MES}: Used to assess marginal risk contribution... & \textit{[Text continues in a single block]} \\
        \vspace{0.1cm}
        \textit{\textbf{Analysis:} SH-PRM uses bullet points to define distinct models, enhancing readability.} & \vspace{0.1cm}
        \textit{\textbf{Analysis:} ORM produces a ``wall of text'' and, in raw outputs, occasionally leaks special tokens (e.g., $<$example$>$) or repeats content, indicating instability.} \\
        \bottomrule
    \end{tabular}
    \caption{\textbf{Comparison of Content Depth and Formatting.} Our model generates professional, well-formatted lists, while the baseline struggles with layout and occasionally exhibits formatting hallucinations.}
    \label{fig:case_formatting}
\end{figure*}

\section{Variance Decomposition}
\label{app:Variance section}
\subsection{Detailed Variance Analysis}
\label{app:Variance Decomposition}
Even if $\mathbb{E}[R_{-t}] = 0$ (unbiased), it introduces significant variance in a single update, which amplifies with $T$. A direct scaling conclusion is:
\begin{equation}
    \mathrm{Var}(R_{-t}) \approx (T-1)\sigma_r^2, \quad \text{where } \sigma_r^2 = \mathrm{Var}(r_t).
\end{equation}

Consequently, from a Signal-to-Noise Ratio (SNR) perspective, the effective learning signal in ORM is diluted by $T$:
\begin{equation}
    \mathrm{SNR}_{\mathrm{ORM}} \propto \frac{\mathrm{Var}(r_t)}{\mathrm{Var}(r_t) + \mathrm{Var}(R_{-t})} \approx \frac{1}{T}
\end{equation}
Thus, the SNR is approximately $O(1/T)$. In contrast, SH-PRM updates the paragraph using only the paragraph-level reward:
\begin{equation}
    \hat{g}_{\mathrm{SH-PRM},t} = \nabla_\theta \log \pi_\theta(a_t \mid s_t) \cdot r_t
\end{equation}
Here, the interference term is zero, and the SNR is approximately $O(1)$. This characterizes the core advantage of SH-PRM: decoupling credit assignment from the ``whole document'' to the ``current paragraph,'' significantly reducing single-step update noise.

\subsubsection{Macro-Consequence: Covariance Explosion}
Microscopically, ORM suffers from the improper inclusion of $R_{-t}$; macroscopically, it leads to strong correlation among gradients within the same trajectory, as the same interference term $R(\tau)$ is shared across all steps $t$. The variance of the total gradient (or batch gradient) can be decomposed into a variance term and a covariance term:
\begin{equation}
    \mathrm{Var}\left( \sum_{t=0}^T \hat{g}_t \right) = \sum_t \mathrm{Var}\left( \hat{g}_t \right) + 2 \sum_{i<j} \mathrm{Cov}\left( \hat{g}_i, \hat{g}_j \right)
\end{equation}

In ORM, the covariance term is often positive due to the shared $R(\tau)$ and accumulates with $T$, causing the overall noise to grow rapidly with sequence length. Using a simplified reward variance decomposition:
\begin{equation}
    \sigma_R^2 = \sum_{t=0}^T \sigma_{r_t}^2 + 2 \sum_{i<j} \mathrm{Cov}(r_i, r_j)
\end{equation}

In long-form generation tasks, paragraph scores often exhibit co-movement (e.g., global readability or insightfulness drives multiple paragraphs to be simultaneously good or bad), making $\mathrm{Cov}(r_i, r_j)$ positive. Consequently, the trajectory variance in ORM can grow as up to $O(T^2)$, whereas SH-PRM's decoupled signals result in significantly weaker covariance and slower noise growth.

Specifically:
\begin{enumerate}
    \item \textbf{Uncorrelated Case:} If $\mathrm{Cov}(r_i, r_j) = 0$ for all $i \neq j$, then $\sigma_R^2 = T\sigma_r^2$ (linear growth).
    \item \textbf{Strongly Correlated Case:} If there exists $\mathrm{Cov}(r_i, r_j) \approx \sigma_r^2$ for all pairs, then $\sigma_R^2 = T^2 \sigma_r^2$ (quadratic growth).
\end{enumerate}

\subsection{Geometric Interpretation: Landscape Smoothing}
\label{app:geometric_interpretation}

From a geometric perspective, the reduction in $V$ and the increase in $F$ fundamentally reshape the ``effective landscape'' traversed by the optimizer, transforming it from a ``rugged plateau'' into a ``smooth funnel.''

The interference term $R_{-t}$ in ORM induces stochastic fluctuations extraneous to the current paragraph, causing the gradient to deviate from the true descent direction. This manifests as a rugged landscape characterized by spurious local minima and optimization instability. By mitigating this interference, SH-PRM aligns the descent direction more consistently, effectively performing \textbf{landscape smoothing} on the objective function, thereby facilitating stable convergence toward stationary points.

This perspective also addresses common skepticism regarding ``statistical equivalence'': while ORM and SH-PRM may share the same set of global optima in the limit of infinite data, their optimization dynamics under finite budgets are \textbf{fundamentally distinct}. In the practical finite-sample regime, the V-F advantage dictates both the quality of reachable solutions and training efficiency.

% \end{document}

% \bibliography{aaai2027}

% \end{document}

\end{document}